\documentclass{article}

\usepackage[preprint]{neurips_2025}

\usepackage[utf8]{inputenc} 
\usepackage[T1]{fontenc}    
\usepackage{hyperref}       
\usepackage{url}            
\usepackage{booktabs}       
\usepackage{amsfonts}       
\usepackage{nicefrac}       
\usepackage{xcolor}         
\usepackage{graphicx}
\usepackage{subcaption}
\usepackage{cite}
\PassOptionsToPackage{hidelinks}{hyperref}
\hypersetup{colorlinks=false, pdfborder={0 0 0}}
\usepackage{float}

\usepackage{todonotes}
\usepackage{siunitx}

\workshoptitle{Machine Learning and the Physical Sciences}

\title{Self-supervised denoising of raw tomography detector data for improved image reconstruction}

\author{%
  Israt Jahan Tulin
  \\
  Department of Information Services and Computing\\
  Helmholtz-Zentrum Dresden-Rossendorf\\
  01328 Dresden, Germany \\
  \texttt{i.tulin@hzdr.de} \\
  \And
  Sebastian Starke \\
  Department of Information Services and Computing\\
  Helmholtz-Zentrum Dresden-Rossendorf\\
  01328 Dresden, Germany \\
  \texttt{s.starke@hzdr.de} \\
  \And
  Dominic Windisch \\
  Institute of Fluid Dynamics\\
  Helmholtz-Zentrum Dresden-Rossendorf\\
  01328 Dresden, Germany \\
  \texttt{d.windisch@hzdr.de} \\
  \And
  André Bieberle \\
  Institute of Fluid Dynamics\\
  Helmholtz-Zentrum Dresden-Rossendorf\\
  01328 Dresden, Germany \\
  \texttt{a.bieberle@hzdr.de} \\
  \And
  Peter Steinbach \\
  Department of Information Services and Computing\\
  Helmholtz-Zentrum Dresden-Rossendorf\\
  01328 Dresden, Germany \\
  \texttt{p.steinbach@hzdr.de} \\
}

\begin{document}

\maketitle

\begin{abstract}
Ultrafast electron beam X-ray computed tomography produces noisy data due to short measurement times, causing reconstruction artifacts and limiting overall image quality. To counteract these issues, two self-supervised deep learning methods for denoising of raw detector data were investigated and compared against a non-learning based denoising method.
We found that the application of the deep-learning-based methods was able to enhance signal-to-noise ratios in the detector data and also led to consistent improvements of the reconstructed images, outperforming the non-learning based method. 
\end{abstract}

\section{Introduction}

Ultrafast electron beam X-ray computed tomography (UFXCT)~[\hyperref[ref:HZDR]{1}] is an innovative, non-invasive imaging technique that has been developed in recent years. It is mainly used for investigations on high dynamic two-phase flows in technical devices, such as chemical reactors, heat exchangers or axial/centrifugal pumps, providing up to 8,000 cross-sectional images per second and a spatial cross-sectional resolution of about 1-2~mm. Reconstruction is performed using filtered back projection~[\hyperref[ref:Windisch2023]{2}]. 

A drawback of the fast imaging rate is the low number of X-ray photons acquired per  projection, i.e. within a sampling interval of 1~$\mu$s. A set of projections that are acquired per rotation of the electron beam are called a sinogram and are used to reconstruct the non-superimposed cross-sectional image. Usage of sinograms with noisy radiation detector values leads to artifacts within the images and reduces their overall quality. Consequently, denoising of UFXCT data is of high interest. Classical algorithms, like applying a Gaussian filter, typically reduce noise effectively, but also reduce image sharpness, resulting in an effective loss of spatial resolution.

The image reconstruction process involves many processing steps which correlate the noise in a non-linear fashion. Changing any parameter in the processing involved may lead to enough of a difference that a trained denoising network can no longer be applied. Therefore, this study aims at denoising the raw sinogram data before involving any image processing steps. Our goal is to capture the systems inherent noise characteristics in a denoising model that can be leveraged for a broad range of applications.

\section{Methods}
\label{sec:methods}


\subsection{Dataset}
\label{ssec:datasets}
A static calibration phantom object was scanned. Data was collected from two detector planes (Plane0 and Plane1) and at three positions (Top, Middle, Bottom) within each plane. For each of these six configurations (combination of plane and position), $2,000$ samples were recorded for a total of $12{,}000$ sinograms with dimensionality $(1000, 144)$ each. Here, the first dimension represents the number of projection angles. The second dimension corresponds to the number of detector channels. \\
In sum, $10{,}000$ samples from five configurations were used for training/validation and the $2{,}000$ samples of the remaining configuration were reserved as a held-out test set. The latter has not been evaluated at the point of writing the manuscript.

\subsection{Model Training, Metrics and Cross-Validation}
\label{ssec:training}
Three popular denoising methods were considered:
1) \textbf{Noise2Void (N2V)}~[\hyperref[ref:N2V]{3}, \hyperref[ref:N2Self]{4}], a self-supervised deep learning method that requires only noisy data for training. N2V masks random pixels during training and predicts their intensity based on the surrounding context, enabling denoising without a clean ground truth,
2) \textbf{N2V2}~[\hyperref[ref:N2V2]{5}], the successor of N2V designed with a focus on reducing checkerboard artifacts and 3) \textbf{Block-Matching and 3D filtering (BM3D)}~[\hyperref[ref:BM3D]{6}], which is a widely used non-learning-based denoising algorithm that exploits patch similarity and collaborative filtering.

Since the dataset lacks true clean sinograms, we created a reference ``ground truth'' for evaluation purposes by averaging across all 2000 samples per fold. This averaging reduces random noise while preserving the underlying structure. Thus, this ground truth serves as a proxy for a clean target.    

We employed a five-fold cross-validation setup on the training data, stratified on detector plane and position. In \textit{sinogram space}, the average of all samples in a fold served as the reference. In \textit{image (reconstruction) space}, the reference was created by averaging the reconstructions of all samples. Image reconstruction was carried out using the RISA~[\hyperref[ref:Windisch2023]{2}] framework by employing fan-beam to parallel-beam conversion, filtering and backprojection.

Denoising performance was quantified using the Peak Signal-to-Noise Ratio (PSNR), calculated between the denoised outputs and their corresponding averaged references.  
Evaluation was performed in both sinogram and reconstruction space.

\begin{figure}[htbp]
    \centering
    \includegraphics[width=0.75\textwidth]{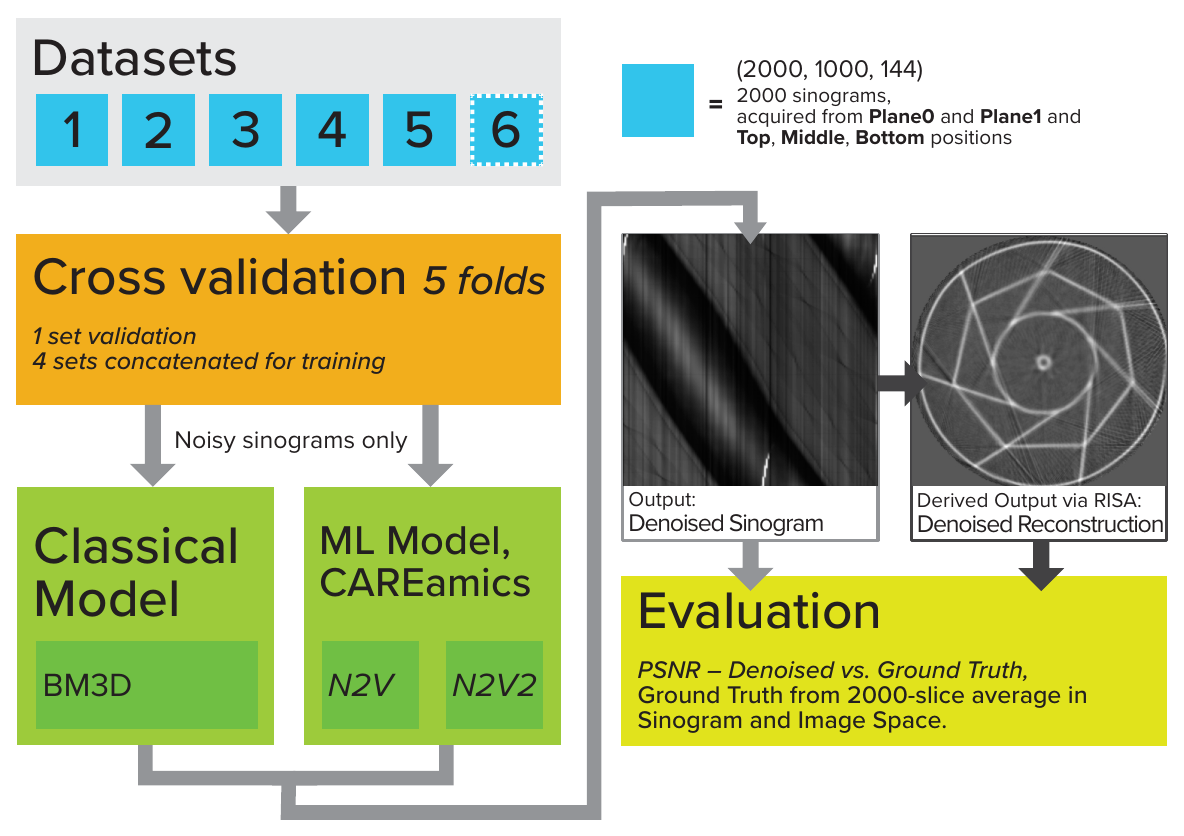}
    \caption{Overview of our training pipeline as detailed in section \ref{sec:methods}. We trained two machine learning based denoising models (N2V and N2V2) using a cross-validation strategy and compared to a non-neural-network-based denoising approach (BM3D).}
    \label{fig:methods_pipeline}
\end{figure}

Both N2V and N2V2 were implemented using the CAREamics framework~[\hyperref[ref:CAREamics]{7}], with training performed only on noisy sinograms since no clean ground truth was available. An overview of the entire pipeline, including dataset organization, cross-validation, models, and evaluation steps, is illustrated in Fig.~\ref{fig:methods_pipeline}.

\subsection{Training Configuration}
Training was carried out on a single node of the local high-performance-computing system using one \textit{NVIDIA A100 GPU} with \qty{40}{GB} of GPU memory. 
All models were trained with the default settings of CAREamics if not otherwise stated. 
We report custom configurations in Appendix~\ref{appendix:training-config}. Our code is available at \url{https://anonymous.4open.science/r/rofex_sinogram_denoising-D4A0}.

\section{Results}

We compared three denoising approaches --- N2V, N2V2, and BM3D --- in both the sinogram space and reconstructed image space. We computed the PSNR value of the original (noisy) data and their denoised counterparts. Performance was measured as the change in PSNR, i.e. the difference in PSNR as $\Delta\text{PSNR}=\text{PSNR}_{denoised} - \text{PSNR}_{original}$, using five-fold cross-validation. We measure the difference in PSNR in units of \unit{dB}.

\begin{figure}[htbp]
    \centering
    \includegraphics[width=\linewidth]{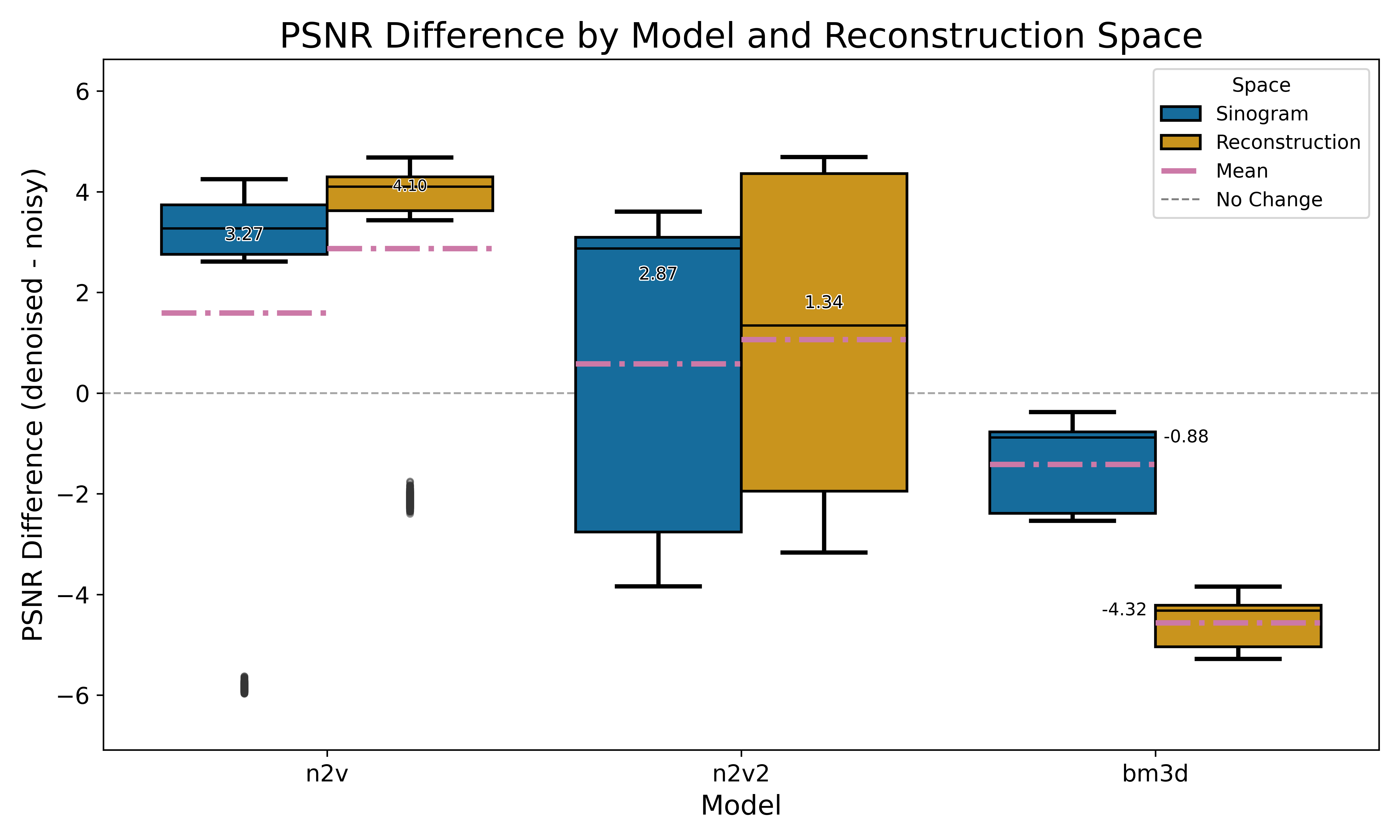}
    \caption{PSNR difference ($\text{PSNR}_{denoised} - \text{PSNR}_{original/noisy}$) for N2V, N2V2, and BM3D across five-fold cross-validation in both sinogram and reconstruction domains. N2V shows stable improvements, N2V2 is more variable, and BM3D consistently underperforms.}
    \label{fig:boxplot}
\end{figure}

As shown in Figure~\ref{fig:boxplot}, N2V gave the most consistent improvements. In the reconstructed image space, it boosted PSNR by about \qty{+4.1}{dB} on median. In sinogram space, the median gains were smaller (\qty{+3.3}{dB}). The tight interquartile ranges indicate that, within both evaluation settings, the denoising performance was stable. A few outliers were observed for N2V, from a single corss-validaiton fold, with poor performance, but this did not affect the overall positive trend.

N2V2 showed more mixed behavior. While its best cases reached similar or even higher improvements than N2V (\qty{+4}{dB}), median PSNR improvements were lower than for N2V, with values of \qty{+2.87}{dB} and \qty{+1.34}{dB} in sinogram and reconstruction space, respectively. Several samples showed little to no improvement, and some even reduced PSNR (down to \qty{-3}{dB} or \qty{4}{dB}). This led to a much wider spread of results, indicating unstable denoising for our use case.

BM3D consistently under-performed, with median PSNR differences of \qty{-0.88}{dB} in the sinogram domain and \qty{-4.32}{dB} in the reconstruction domain, indicating that this non-learning-based method is not well suited to the noise present in our data.

To better illustrate these findings, we examined reconstructed samples from the best-performing fold of N2V2 and N2V, using the same input for both methods. Representative examples are provided in Appendix~\ref{appendix:denoising-n2v2} and Appendix~\ref{appendix:denoising-n2v}. For comparison, we also include BM3D results in Appendix~\ref{appendix:denoising-bm3d}. Both N2V and N2V2 reduced background noise while preserving structural details. BM3D produced visible smoothing but also removed or blurred fine structural details (e.g., tiny textures, edges, and sharp transitions). 

We also analyzed the difference between noisy and denoised images to better understand the model’s behavior (see Appendix~\ref{appendix:noisy-denoised}). In well-performing examples (top row), the noise map looks mostly like random speckle and does not reproduce the object’s structure. This implies the model mainly removed noise while keeping the real features intact. In an under-performing example (bottom row), the noise map shows shapes of the scanned object (rings and spokes), indicating that the model also removed true signal, not just noise. We consider this to be an over-smoothing effect. Table~\ref{tab:noise_stats} summarizes the noise statistics for a good and a bad fold. The bad fold shows substantially larger mean differences and extreme values, consistent with the over-smoothing visible in Fig.~\ref{fig:good_bad_fold}.


\section{Discussion and Limitations}
In summary, our results demonstrate that deep-learning-based denoising methods (N2V and N2V2) outperformed the classical baseline (BM3D) on sinogram data, highlighting the advantage of self-supervised approaches for this task on our data set.

Although assuming independent noise, N2V still produced stable improvements across folds, resulting in reliable performance in both sinogram and reconstruction spaces. N2V2, in contrast, was designed to address checkerboard artifacts introduced by N2V’s masking and pooling steps—artifacts that were not a major issue in our data. In our case, detector streaks and ring-like structures (generated by the fan-beam to parallel-beam interpolation and the subsequent inverse Radon-transform during backprojection) were more prominent. N2V2's architectural changes (removing a skip connection and replacing max pooling with blur pooling) likely reduced the network’s ability to preserve fine details. Blur pooling also introduces extra smoothing, which might remove useful high-frequency information together with noise. As a result, N2V2 appeared more sensitive to the specific characteristics of our data, which may explain its less stable performance across folds. 

BM3D, used as a classical baseline, consistently under-performed and showed a tendency to reduce PSNR. This suggests that this traditional patch-based filtering approach, designed with natural imaging data in mind, is not well-suited to our highly structured noise of CT detector signals.

Interestingly, performance in the reconstruction space was generally better than in the sinogram space for both deep learning approaches. We believe this to be caused by the reconstruction which combines information across many projections. This in turn reduces residual noise and amplifies the benefits of denoising.

Several limitations and directions for future work also became apparent in this study. 
First, while our results show consistent PSNR gains, it remains unclear whether these improvements translate into practical benefits compared to simply averaging reconstructions. PSNR may also not reflect actual image quality for all scanned structures. We therefore plan to evaluate the impact of denoising in a downstream application of practical relevance as a next step.  
Second, although N2V2 showed promise in certain folds, its instability limits its practical use. Further tuning of its hyperparameters could be explored to enhance performance. Third, autocorrelation analysis (see Appendix~\ref{appendix:autocorrelation}) revealed strong diagonal correlations in the sinogram data. 
While CAREamics provides \emph{StructN2V} for handling structured noise (via horizontal or vertical masking), it does not yet support diagonal masking. 
Finally, this study was based on data from a single phantom. To fully characterize noise properties and train models that generalize better, more diverse datasets are required. Such diversity might allow the networks to disentangle noise from object structure more effectively.

In sum, self-supervised methods (N2V/N2V2) outperformed BM3D for denoising CT sinogram data, with N2V showing the most stable performance. We found that denoising of raw detector data also led to improved image reconstructions, motivating further work and validation on broader datasets and downstream tasks.




\section*{References}


{
\small

\noindent\label{ref:HZDR}
[1] F. Fischer, U. Hampel
Ultra fast electron beam X-ray computed tomography for two-phase flow measurement
Nuclear Engineering and Design 240(9), 2254-59, 2010.

\medskip

\noindent\label{ref:Windisch2023}
[2] Dominic Windisch, Jeffrey Kelling, Guido Juckeland, and André Bieberle.  
“Real-time data processing for ultrafast X-ray computed tomography using modular CUDA-based pipelines.”  
In \textit{Computer Physics Communications}, vol. 287, article 108719, 2023.  
\url{https://doi.org/10.1016/j.cpc.2023.108719}

\medskip

\noindent\label{ref:N2V}
[3] Alexander Krull, Tim-Oliver Buchholz, and Florian Jug.
“Noise2Void: Learning denoising from single noisy images.”
In \textit{Proceedings of the IEEE/CVF Conference on Computer Vision and Pattern Recognition (CVPR)}, 2019.
\url{https://openaccess.thecvf.com/content_CVPR_2019/html/Krull_Noise2Void_-_Learning_Denoising_From_Single_Noisy_Images_CVPR_2019_paper.html}

\medskip

\noindent\label{ref:N2Self}
[4] Joshua Batson, and Loic Royer.
“Noise2Self: Blind denoising by self-supervision.”
In \textit{Proceedings of the 36th International Conference on Machine Learning (ICML)}, PMLR 97:524–533, 2019.
\url{https://proceedings.mlr.press/v97/batson19a.html}

\medskip

\noindent\label{ref:N2V2}
[5] Eva Höck, Tim-Oliver Buchholz, Anselm Brachmann, Florian Jug, and Alexander Freytag.  
“N2V2 – Fixing Noise2Void Checkerboard Artifacts with Modified Sampling Strategies and a Tweaked Network Architecture.”  
In \textit{Computer Vision – ECCV 2022 Workshops}, Lecture Notes in Computer Science, vol 13804. Springer, Cham, 2023.  
\url{https://doi.org/10.1007/978-3-031-25069-9_33}

\medskip

\noindent\label{ref:BM3D} [6] K. Dabov, A. Foi, V. Katkovnik, and K. Egiazarian, 
“Image denoising by sparse 3-D transform-domain collaborative filtering,” 
\textit{IEEE Trans. Image Process.}, vol. 16, no. 8, pp. 2080--2095, Aug. 2007, 
doi: 10.1109/TIP.2007.901238.
\url{https://ieeexplore.ieee.org/document/4271520}

\medskip

\noindent\label{ref:CAREamics}
[7] CAREamics: A PyTorch library for Noise2Void and related denoising methods.  
Available at: \url{https://careamics.github.io/0.1/}

}


\appendix

\section{Additional Details}

\subsection{Training Configuration}\label{appendix:training-config}

We employed the CAREamics framework~[\hyperref[ref:CAREamics]{7}] to denoise raw sinograms of the ROFEX detector. We deviated from the default values in version \texttt{0.0.15} as documented below:

\begin{itemize}
    \item \textbf{Training parameters}:
    \begin{itemize}
        \item Patch size: $64 \times 64$  
        \item Batch size: 64  
        \item Number of epochs: 200  
        \item Data loader workers: 4 (training), 2 (validation)
    \end{itemize}
    \item \textbf{Denoising configuration}:
    \begin{itemize}
        \item ROI size: 7  
        \item Masked pixel percentage: 0.5  
        \item For N2V2, \texttt{use\_n2v2=True}; for N2V, \texttt{use\_n2v2=False}
    \end{itemize}
\end{itemize}

All other hyperparameters, such as optimizer type, learning rate scheduler, and network architecture, followed CAREamics default values.

\subsection{Denoising performance with N2V2}\label{appendix:denoising-n2v2}

\begin{figure}[H]
    \centering
    \includegraphics[width=\textwidth]{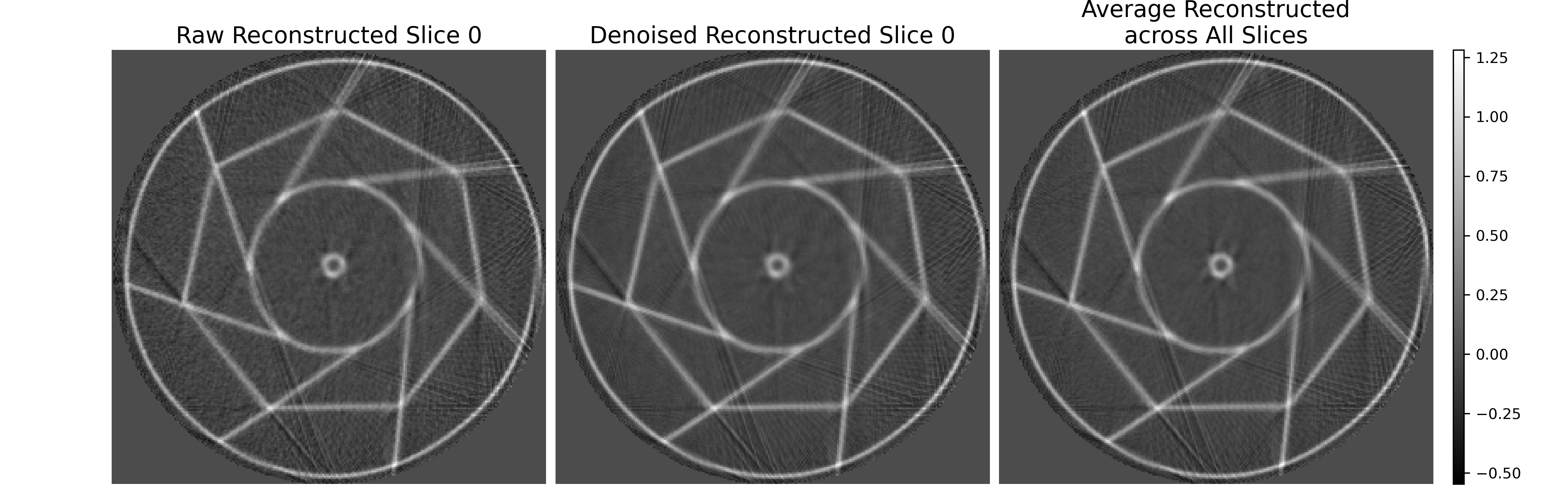}
    \caption{Example reconstructed slice from the best-performing fold of N2V2, showing effective noise suppression while preserving structural details.}
    \label{fig:reconstructions_n2v2}
\end{figure}

\subsection{Denoising performance with N2V}\label{appendix:denoising-n2v}

\begin{figure}[H]
    \centering
    \includegraphics[width=\textwidth]{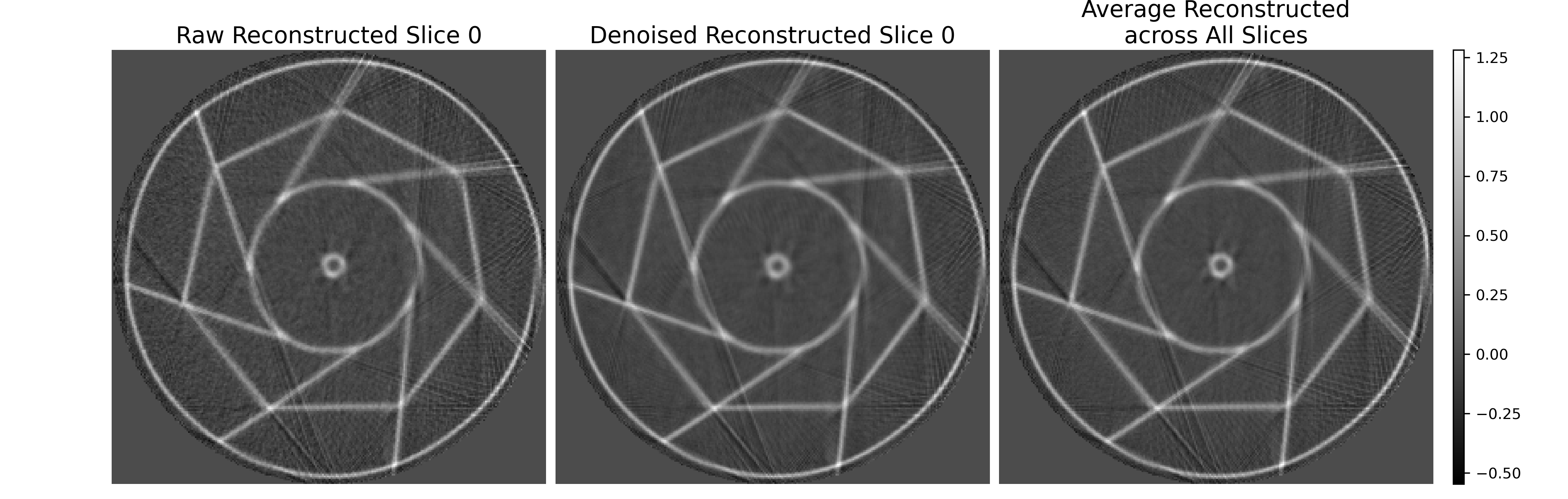}
    \caption{Example reconstructed slice from the best-performing fold of N2V, demonstrating consistent denoising and structural preservation.}
    \label{fig:reconstructions_n2v}
\end{figure}

\subsection{Denoising performance with BM3D}\label{appendix:denoising-bm3d}

\begin{figure}[H]
    \centering
    \includegraphics[width=\textwidth]{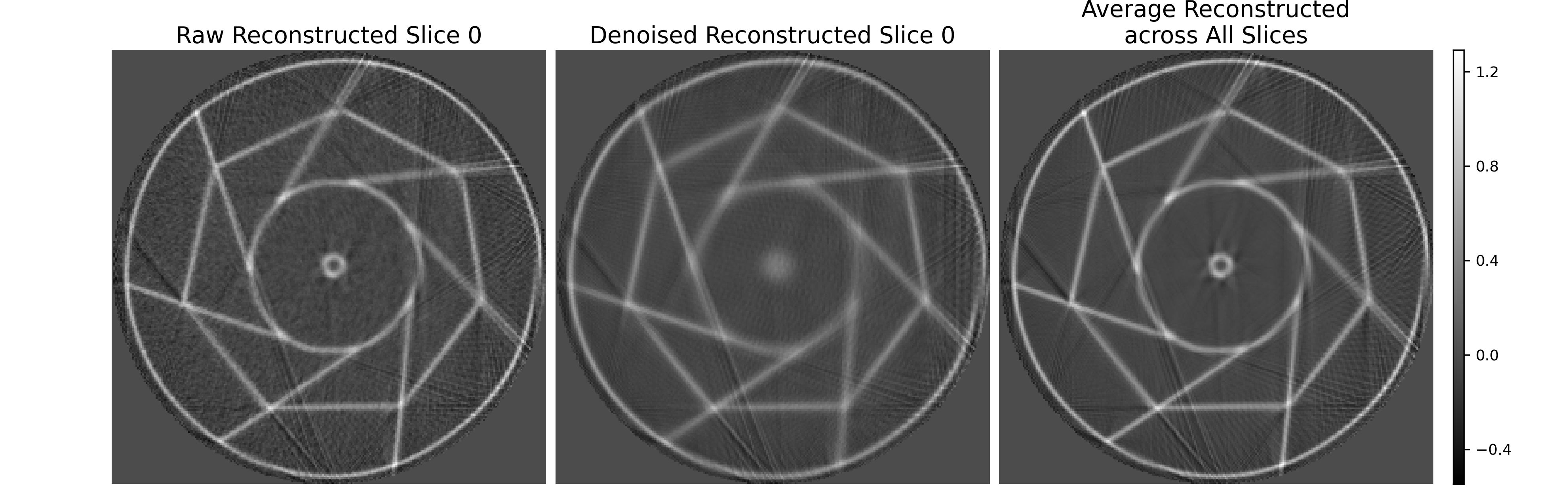}
    \caption{Example reconstructed slice from BM3D denoising, illustrating its filtering effect on the noisy input.}
    \label{fig:reconstructions_bm3d}
\end{figure}

\subsection{Difference between noisy and denoised images}\label{appendix:noisy-denoised}

\begin{figure}[H]
    \centering
    \includegraphics[width=\textwidth]{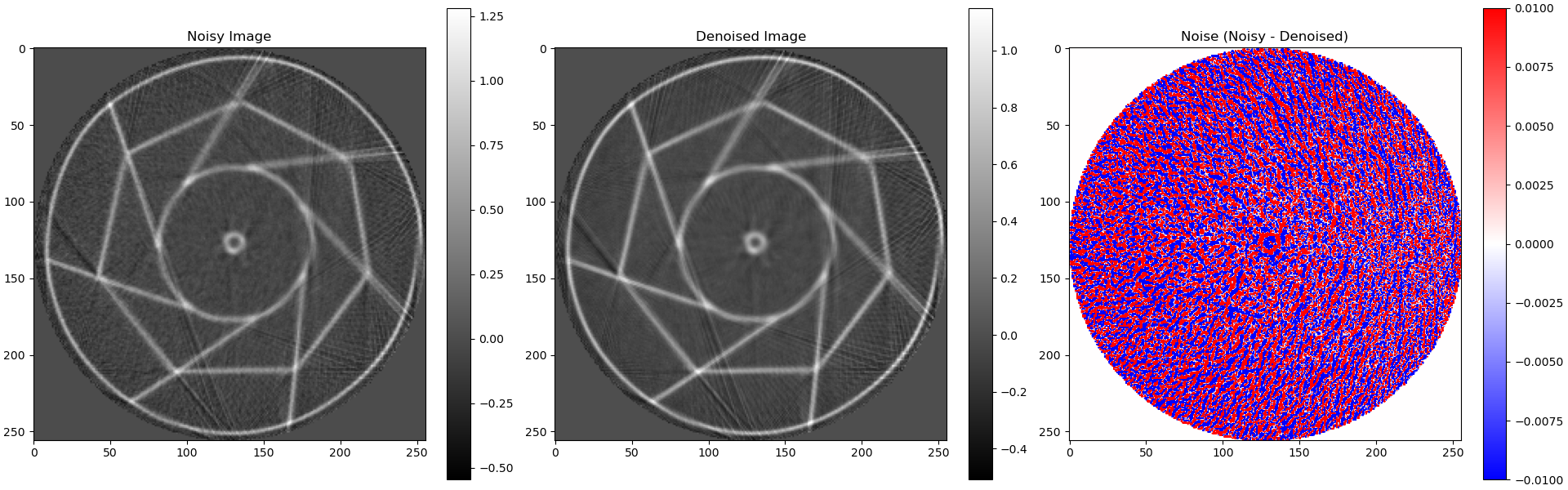}
    \includegraphics[width=\textwidth]{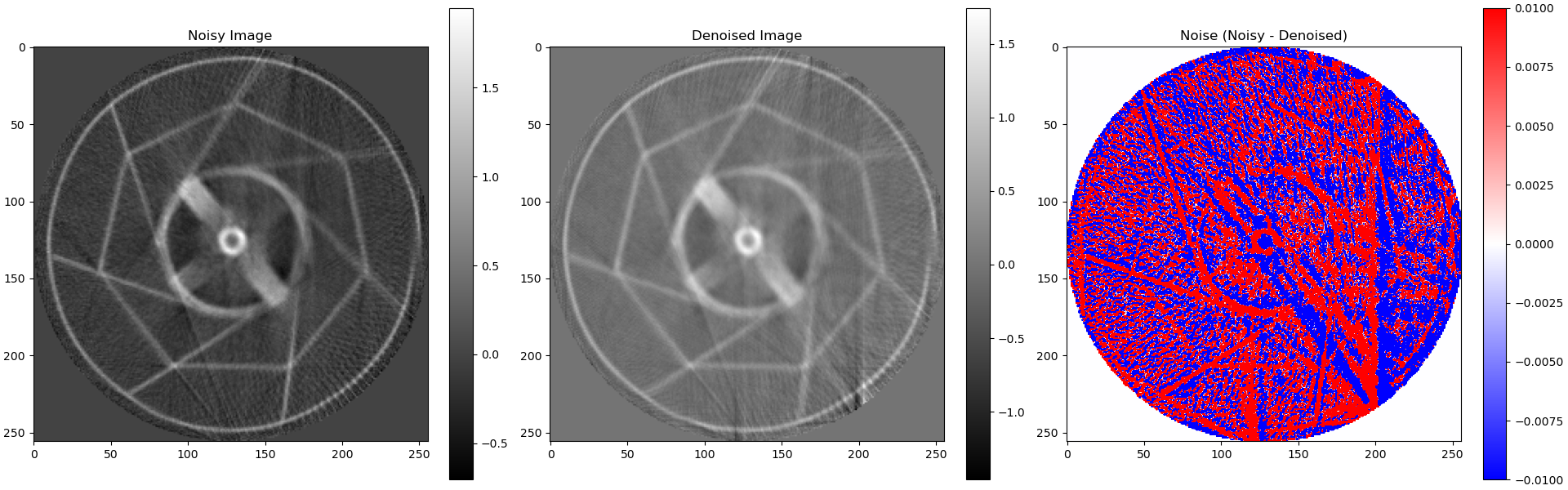}
    \caption{Qualitative comparison of denoising performance across folds. 
    (Left) Noisy reconstruction. 
    (Middle) Denoised reconstruction with N2V2. 
    (Right) Noise image, computed as \textit{noisy -- denoised}. 
    The noise maps are shown in a blue–red diverging colormap, where red indicates positive differences and blue indicates negative differences. 
    \textbf{Top:} Example of a good fold, where the noise map shows mostly random residuals and no clear object structure. 
    \textbf{Bottom:} Example of a bad fold, where the noise map reveals structured residuals corresponding to true features, indicating that signal was also removed.}
    \label{fig:good_bad_fold}
\end{figure}

\begin{table}[htbp]
  \centering
  \caption{Noise statistics (\(\text{noisy}-\text{denoised}\)) for N2V2 on a good vs.\ bad fold. Values are computed on reconstructed slices.}
  \label{tab:noise_stats}
  \begin{tabular}{lrrrr}
    \toprule
    Fold & Mean abs & Std & Max & Min \\
    \midrule
    Good & 0.026 & 0.038 & 0.392 & -0.401 \\
    Bad  & 0.057 & 0.087 & 1.501 & -1.475 \\
    \bottomrule
  \end{tabular}
\end{table}

\subsection{Analysis of correlated noise}\label{appendix:autocorrelation}
\begin{figure}[H]
    \centering
    \includegraphics[width=\linewidth]{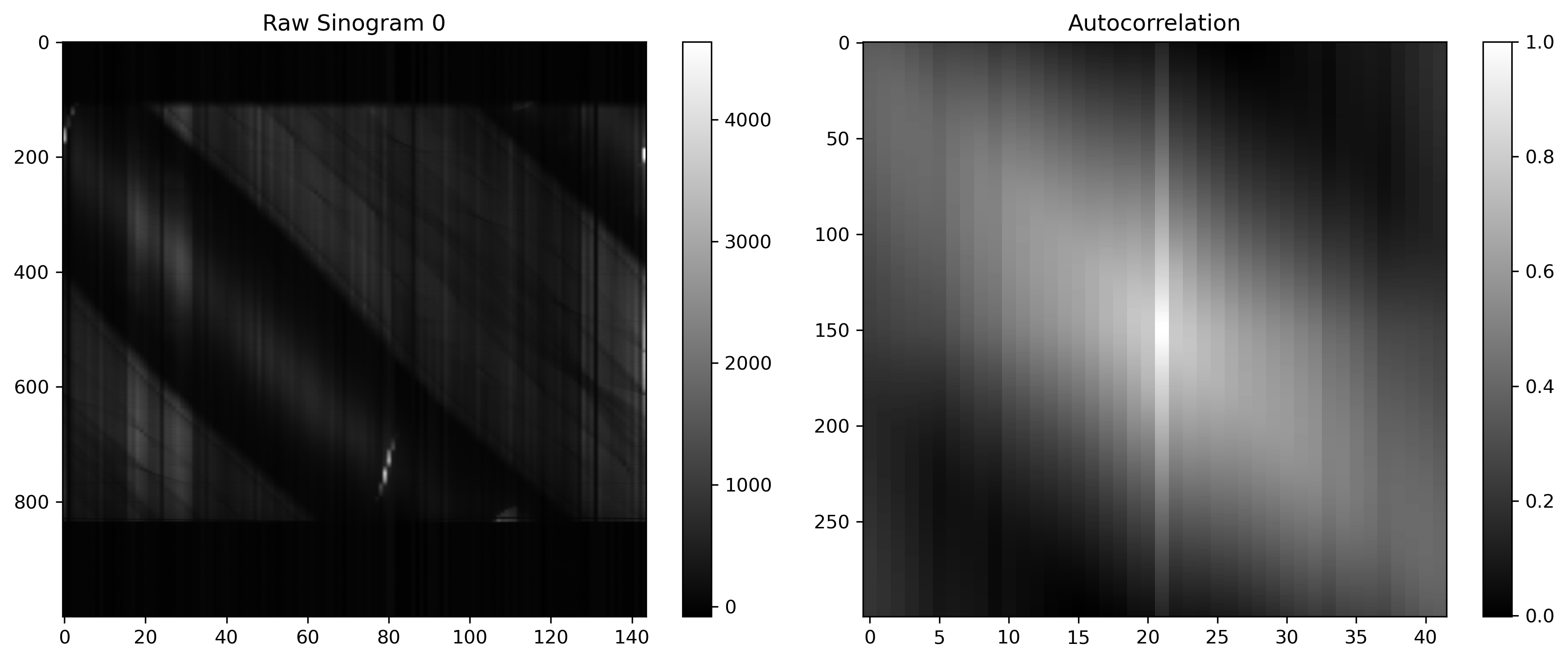}
    \caption{Autocorrelation analysis of raw sinogram data. 
    (Left) Example of a raw sinogram containing structured artifacts.
    (Right) Corresponding autocorrelation map, showing extended diagonal correlations that indicate structured noise across both detector bins and projection angles.}
    \label{fig:autocorr}
\end{figure}

\end{document}